\documentclass[twocolumn,10pt]{IEEEtran}

\usepackage{verbatim}
\usepackage{amsfonts}
\usepackage{amssymb}
\usepackage{stfloats}
\usepackage{cite}
\usepackage{graphicx}
\usepackage{psfrag}
\usepackage{subfigure}
\usepackage{amsmath}
\usepackage{array}
\usepackage{epstopdf}
\usepackage{authblk}
\usepackage{graphicx} 
\usepackage{amsthm} 
\usepackage{lipsum}
\usepackage{verbatim} 
\usepackage{authblk}
\usepackage{mathtools}
\usepackage{cuted}
\usepackage[lined,boxed,ruled]{algorithm2e}
\usepackage{booktabs}

\usepackage[usenames]{color}

\newtheorem{remark}{\bf Remark}

\def\phi{\varphi}

\def\({\left(}
\def\){\right)}

\def\b0{{\mathbf{0}}}

\makeatletter
\newcommand{\removelatexerror}{\let\@latex@error\@gobble}
\makeatother

\begin{document}

\graphicspath{{figure/}}

\title{\huge Federated Dropout -- A Simple Approach for Enabling Federated Learning on Resource Constrained Devices}
\author{Dingzhu Wen, Ki-Jun Jeon, and Kaibin Huang     \thanks{\setlength{\baselineskip}{13pt} \noindent D. Wen (e-mail: wendzh@shanghaitech.edu.cn) is with the School of Information Science and Technology in ShanghaiTech University, China. K.-J. Jeon (e-mail: kj.jeon@lge.com) is with the LG Electronics, Korea. K. Huang (e-mail: huangkb@eee.hku.hk)  is with  The  University of  Hong Kong, Hong Kong. Corresponding author: Kaibin Huang. }
}

\maketitle

\begin{abstract}
\emph{Federated learning} (FL) is a popular framework for training an AI model using distributed mobile data in a wireless network. It features  data parallelism by distributing the learning task to multiple edge devices while attempting to preserve  their local-data privacy.  One main challenge confronting  practical FL is that resource constrained devices struggle with the computation  intensive task of updating  of a deep-neural network model.  To tackle the challenge, in this paper, a \emph{federated dropout} (FedDrop) scheme is proposed building on the classic dropout scheme for random model pruning. Specifically, in each  iteration of the FL algorithm, several subnets are independently generated from the global model at the server using dropout but with heterogeneous  dropout rates (i.e., parameter-pruning probabilities), each of which is adapted to the state of an assigned  channel. The subnets are downloaded to associated  devices for updating. Thereby, FedDrop reduces both the communication overhead and devices' computation loads compared with the conventional  FL while outperforming  the latter in the case of overfitting   and also the FL scheme with uniform  dropout (i.e., identical subnets). 

\end{abstract}

\section{Introduction}
The existence of enormous  data at edge devices (e.g., sensors and smartphones) is driving the deployment of machine-learning  algorithms at the network edge, called \emph{edge learning},  to distill the data into intelligence, which supports  AI  powered  applications ranging from virtual reality to eCommerce \cite{letaief2019roadmap,lin2020edge}.  The  \emph{federated learning} (FL)  is perhaps the most popular framework mainly due to its feature of leveraging mobile data while preserving their privacy \cite{wang2018edge,amiri2020machine}. Based on the  \emph{stochastic gradient descent} (SGD) algorithm, FL distributes the task of training an AI model at multiple devices. The algorithm requires devices to periodically upload to a server high-dimensional models (or stochastic gradients) that are  computed using local data and used to  update the global model after aggregation. To be specific, a typical  AI model (e.g.,  AlexNet or  ResNet50) has tens of millions of  parameters. As a result, the deployment of FL in a wireless network faces two main challenges: 1) overcoming the \emph{communication bottleneck} due to finite radio resources, and 2) overcoming the \emph{computation bottleneck} due to finite computation capacities of devices.

One main research focus in the area of FL is to tackle the communication bottleneck. To improve the communication efficiency, researchers have proposed new communication techniques including over-the-air aggregation   \cite{zhu2019broadband, ShiyuanmingAirComp, amiri2020machine}, joint computation-and-communication  resource management \cite{chen2020joint, ren2020scheduling, shi2020device,yang2019scheduling}, and high-dimensional source encoding \cite{du2020high}. Most of these techniques do not address the issue of computation bottleneck as they still require devices to update the full models. On the other hand, there exist approaches of coping with the  bottleneck including reducing the sizes of data mini-batches, tolerating long learning latency or excluding ``stragglers" that are slow in computing. Such approaches attempt to solve the problem at the cost of performance degradation. Recently, researchers have explored a potentially more effective approach of partitioning the model and allocating only a part of the model  to each device for updating. The resultant framework, called \emph{partitioned edge learning}, can overcome the computation bottleneck without incurring any performance loss if the loss function, which is the learning objective, is linearly decomposable (e.g., support vector machine) \cite{wen2020joint,wen2020adaptive}. However, this property does not hold for the popular \emph{deep neural networks} (DNNs). Attempts have been made on making DNNs decomposable by introducing auxiliary variables \cite{carreira2014distributed}. This technique, however, not only increases the computation complexity but also requires an additional communication round to complete the original single communication round of  the conventional FL algorithm without model partitioning. 

In this paper, we propose a novel approach of simultaneously tackling   both the \emph{communication-and-computation} ($C^2$) bottlenecks, termed \emph{federated dropout} (FedDrop), to enable federated learning on resource-constrained devices. The approach is inspired by  the well-known   dropout technique that was originally  developed in computer science for preventing overfitting of a DNN model \cite{srivastava2014dropout}. To this end, the  neurons (together with their connections)  in the  DNN  are  randomly dropped during training.  Each instance of dropout reduces the model to a reduced-dimension version, called \emph{subnet}. The model-training process involves updating a random sequence of subnets. Acting as an adaptive regularization mechanism, dropout  prevents DNN parameters from excessive  co-adapting and as a result avoids  model  overfitting, thereby improving the performance of different  supervised learning tasks, such as vision and speech recognition  \cite{wager2013dropout}. While the main purpose of dropout is to solve the overfitting problem, FedDrop is more sophisticated and aims at overcoming the $C^2$ bottlenecks in FL. For the purpose, FedDrop is designed to have three distinctive  features: 
\begin{itemize}
\item \emph{Subnet distribution} --  A server assigns  randomly generated subnets to different devices for updating. As devices are now required to train and transmit subnets instead of full models, the  $C^2$ bottlenecks are alleviated.

\item \emph{Subnet aggregation} -- The server aggregates  updated subnets for updating the global model. 

\item \emph{$C^2$ awareness} --  Define the dropout rate of a device as  the probability of deactivating a neuron of the model in the generation of the assigned subnet. The  dropout rates of the devices are  adapted to their  $C^2$ states (i.e., communication rate and computation capacity) so as to reduce the learning latency.

\end{itemize}

Compared with the naive approach of updating a single subnet by all devices in each round, called \emph{uniform dropout}, FedDrop updates the whole model and thereby achieves better learning performance as observed in experiments (e.g., an accuracy increase of $2.5\%$ under the dropout rate of $0.6$ for training the dataset Cifar-10). Specifically, in the proposed FedDrop scheme,  the server uses the conventional dropout technique to randomly  generate multiple subnets, each of which is assigned to a single device for updating. The dropout rate of each device   is adapted to the assigned device's $C^2$ state under a per-round latency constraint for  FL with  synchronized updating.  Let  $p_k$ denote the dropout rate of device $k$. For the FedDrop schemes, it can be shown that the $C^2$ loads  of  each device  can be reduced by  a ratio of $(1 - p_k)^2$ for device $k$. As shown by experimental results, with dropout rates up to $60\%$, FedDrop  can improve the learning performance in the case of training a large-scale  model  (with $74,008,736$ parameters) on a complex dataset (e.g., Cifar) by avoiding overfitting. In the case of relatively small-scale model (with $17,250$ parameters) on a simpler dataset (e.g., MNIST), FedDrop incurs only slight performance degradation (e.g., an accuracy decrease of $0.88\%$ under the dropout rate of $0.3$).

%The reminder of this paper is organized as follows. Section II introduces the system model. The FL with FedDrop scheme is proposed in Section III. The experiments and performance analysis are presented in Section IV and Section V concludes the paper.

\section{System Model}
\begin{figure*}[t]
\centering
\includegraphics[width=0.8\textwidth]{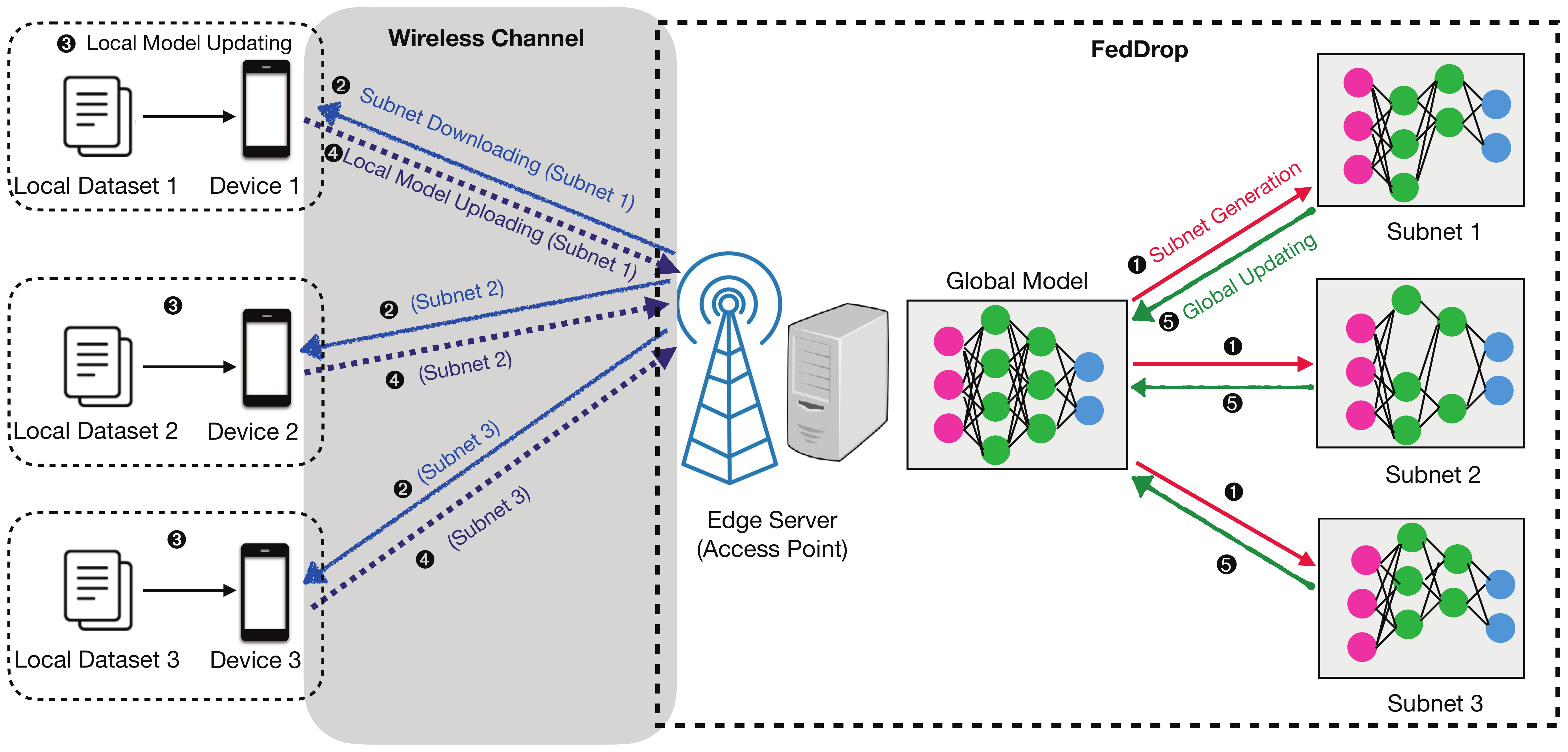}
\caption{The operations of FL with FedDrop in a wireless system.}\label{Fig:SysModel}
\end{figure*}

Consider a single-cell FedDrop system, as shown in Fig. \ref{Fig:SysModel}, where there are one server equipped with an access point and $K$ edge devices targeting cooperatively finishing a FL task with FedDrop. The devices are connected to the server via wireless links. In each training iteration (called \emph{one communication round}), each device uses an orthogonal bandwidth for downloading and uploading. Frequency non-selective channels are considered. The server works as a coordinator and has the channel state information of all links. 

\subsubsection{FL}
In a traditional FL system, the training of a DNN model iterates among three steps: 1) a server broadcasts a global model to all devices; 2) Each device calculates a local update based on its local dataset; 3) the server aggregates all local updates for global updating. In this paper, the scheme of FedDrop is used in FL to reduce the $C^2$ overhead of each device. To this end, new training strategy with additional steps in each round is required, as shown in Fig. \ref{Fig:SysModel}, where the server randomly generates several subnets using the dropout technique with each being assigned to one device for updating. The design of the FL with FedDrop scheme is elaborated in Section \ref{Sect:FedDrop}.

\subsubsection{Dropout}
To facilitate the discussion of FedDrop, the original dropout scheme as presented in \cite{srivastava2014dropout} is introduced as follows. First of all, dropout is  only applied to the fully connected layers in a DNN (or a fully connected neural network) since other layers (e.g., convolutional layers) have much fewer parameters. Consider a fully connected layer in a DNN. The  output of the $i$-th neuron in the $\ell$-th layer is denoted as $f_{\ell,i}({\bf w}_{\ell,i})$ where $f_{\ell,i}(\cdot)$ is  the activation function [e.g., \emph{rectified linear unit} (ReLU)], and ${\bf w}_{\ell,i}$ is the parameter vector. The dropout technique generates  a subnet by deactivating each neuron (together with its connections) with a probability of $p_k$ in each round of model training. The probability  $p_k$ is called  the \emph{dropout rate} and $(1-p_k)$ the \emph{presence rate}. Specifically, with random dropout, an arbitrary neuron, say the $i$-th neuron in the $\ell$-th layer, can be expressed as a random variable: 
\begin{equation}
    \hat{f}_{\ell,i}({\bf w}_{\ell,i}) =m_{\ell,i}^{(k)} f_{\ell,i}({\bf w}_{\ell,i}), \quad \forall (l,i),
\end{equation}
where $m_{\ell,i}^{(k)}$ is the dropout mask, defined as
\begin{equation}\label{Eq:Mask}
m_{\ell,i}^{(k)} = \left\{
    \begin{aligned}
    &\dfrac{1}{1 - p_k}, && \text{w.p. } (1 - p_k),\\
    & 0, && \text{w.p. } p_k.
    \end{aligned}
    \right.
\end{equation}
 In \eqref{Eq:Mask}, the scaling factor $1/(1-p_k)$ guarantees that the expectation of $\hat{f}_{\ell,i}({\bf w}_{\ell,i})$ is $f({\bf w}_{\ell,i})$. As a result,  the expectation of all neurons' outputs in the subnet is equal to those in the original DNN.
 
As mentioned, in a subnet generated by dropout, each neuron in the subnet is randomly deactivated with a probability of $p_k$. In this case, the size of the subnet is random and  the $C^2$ overhead to train it  can not be determined. This further leads to uncertain learning latency. To address issue, an approach of \emph{progressive random parametric pruning} is adopted for realizing the dropout scheme to generate a subnet with deterministic size, as descried below.
\begin{itemize}
\item \emph{Step 1}: For an arbitrary fully connected layer in the DNN, randomly select a neuron with a uniform probability and deactivate it.
\item \emph{Step 2}: Repeat Step 1 until  the dropout rate is satisfied, i.e., exactly $p_k$ of the neurons in the layer are deactivated. 
\end{itemize}

\subsubsection{Per-Round Latency}
Consider an arbitrary round and an arbitrary device, say the $k$-th device. It is assigned by the server with a subnet, which is generated using the dropout technique with a dropout rate of $p_k$. Then, the latency for device $k$ to update its subnet in this round includes two parts. One is the communication latency for downloading the subnet and uploading the corresponding updates, given by 
\begin{equation}\label{Eq:COMLatency}
T^{\rm com}_k  = \dfrac{M_k Q}{ B_kR_k^{ d\ell}} +  \dfrac{M_k Q}{ B_k R_k^{ u\ell}}, \quad 1\leq k\leq K,
\end{equation}
where the first term is the downloading latency, the second term is the uploading latency, $M_k$ is the number of parameters of the subnet and will be analyzed later [see \eqref{Eq:Parameters}], $Q$ is the quantization bits used for one parameter, $B_k$ is the bandwidth assigned to device $k$, and $R_k^{d\ell}$ and $R_k^{u\ell}$ are the downlink and uplink spectrum efficiencies of device $k$, respectively. The other is the computation latency for local calculation. According to \cite{you2016energy}, it is given by
\begin{equation}\label{Eq:CMPLatency}
T^{\rm cmp}_k = \dfrac{C_k D_k}{f_k}, \quad 1\leq k \leq K,
\end{equation}
where $C_k$ is the number of processor operations required for updating the subnet regarding one data sample and  will be analyzed later [see \eqref{Eq:CMPLoad}], $D_k$ is the number of samples trained on device $k$, and $f_k$ is the computation speed of device $k$. Overall, the latency of device $k$ in this round is
\begin{equation}\label{Eq:DeviceLatency}
    T_k = T^{\rm com}_k + T^{\rm cmp}_k, \quad 1\leq k \leq K,
\end{equation}
where  $T^{\rm com}_k$ is the communication latency defined in \eqref{Eq:COMLatency} and $T^{\rm cmp}_k$ is the computation latency defined in \eqref{Eq:CMPLatency}. Finally, due to the \emph{synchronized updates} property (see e.g., \cite{wen2020joint}), i.e., the server can not finish one training round before all devices upload their updates, the latency of this round depends on the slowest device:
\begin{equation}\label{Eq:SystemLatency}
    T = \max\limits_k\; T_k.
\end{equation}

\section{Federated Dropout}

\subsection{Federated Learning  with FedDrop}\label{Sect:FedDrop} 
The scheme  of federated  learning scheme enhanced by FedDrop is described as follows. The feature of FedDrop is to generate subnets of different sizes for distributed updating at devices such that the sizes are adapted to their $C^2$ capacities. In each training round of FL with FedDrop, the operations are shown in Fig. \ref{Fig:SysModel} and are elaborated below.
\begin{enumerate}
\item \emph{Subnets generation}: For each device, the server generates a subnet, say $\mathcal{S}_k$ for the $k$-th device, with the FedDrop rate defined in \eqref{Eq:MimimumRates}.

\item \emph{Model downloading}: Each device downloads its assigned subnet from the server.

\item \emph{Local model updating}: All devices update their corresponding local subnets based on their local datasets.

\item \emph{Local model uploading}: Each device uploads its updated local model to the server.

\item \emph{Global model updating}: In the server,  for an arbitrary subnet, say $\mathcal{S}_k$, a complete DNN, say $\mathcal{N}_k$, is first constructed, where the parameters included in $\mathcal{S}_k$ are updated and the other parameters use the same values in the last round. Then, the global model is updated by averaging all complete DNNs, say $\{\mathcal{N}_k\}$.
\end{enumerate} 

\begin{remark}[FL versus FL with FedDrop]
In FL, a global model is used in each round and broadcast is used for model downloading. In FL with FedDrop, each device can adaptively select a subnet for updating based on its $C^2$ capacities.
\end{remark}

%The convergence of the above algorithm can be analyzed by relatively straightforward  modification and  combination of existing analysis on dropout and stochastic gradient descent \cite{yuan2019distributed,carter1991global}. The details are omitted due to the space limitation.

\subsection{Analysis and Optimization of FedDrop}
In each training round of the FedDrop scheme, devices are assigned by a server with different subnets generated via different dropout rates. In the sequel, we first analyze the  $C^2$ overhead of each device. Then, the adaptive FedDrop rate of each device is designed based on given per-round latency $T$, varying spectrum efficiencies $\{R_k^{ d\ell},R_k^{ u\ell}\}$, and the allocated bandwidth $\{B_k\}$. 

\subsubsection{Communication overhead analysis}
Consider an arbitrary device, say the $k$-th. Its communication overhead in each round depends on the number of parameters of its assigned subnet. First, consider any two adjacent fully connected layers in the original DNN, say the $\ell$-th layer with $N_{\ell}$ neurons and the $(\ell+1)$-th layer with $N_{\ell+1}$ neurons. The number of parameters to connect these two layers is $N_{\ell}\times N_{\ell+1}$. Then, the dropout technique is used to randomly generate a subnet for the device with a dropout rate of $p_k$. In the subnet, the numbers of neurons in the two layers can be derived as $(1-p_k)N_{\ell}$ and $(1-p_k)N_{\ell+1}$, respectively. Furthermore, the number of parameters to connect these two layers in the subnet can be derived as $(1-p_k)^2N_{\ell}\times N_{\ell+1}$. Next, as the dropout technique is applied in all fully connected layers, it's easy to derive the number of parameters in the fully connected layers of the subnet as $(1-p_k)^2 M_{\rm full}$, where $M_{\rm full}$ is the total number of parameters in these layers. Hence, the total number of parameters in the subnet is given by
\begin{equation}\label{Eq:Parameters}
M_k = M_{\rm conv} + (1-p_k)^2 M_{\rm full}, \quad 1\leq k \leq K,
\end{equation}
where $M_{\rm conv}$ and $M_{\rm full}$ are the numbers of parameters in the convolutional layers and fully connected layers of the original DNN, respectively.

\subsubsection{Computation overhead analysis}
Similarly, consider the $k$-th device, the $\ell$-th and the $(\ell+1)$-th layers in the original DNN. The computation between these two layers have two phases. One is the forward phase, whose operations contain $N_{\ell}\times N_{\ell+1}$ multiplications, $(N_{\ell}-1)\times N_{\ell+1}$ additions, and $N_{\ell+1}$ activation operations. The other is the backward phase, which calculates the gradients of $N_{\ell}\times N_{\ell+1}$ parameters. Then, a subnet is randomly generated for the device with a dropout rate of $p_k$. In the forward phase, the numbers of multiplications, additions, and activation operations to update the subnet are $(1-p_k)^2 N_{\ell}\times N_{\ell+1}$, $[(1-p_k) N_{\ell}-1]\times (1-p_k) N_{\ell+1}$, and $(1-p_k) N_{\ell+1}$, respectively. In the backward phase, the number of gradients to be calculated in the subet is $(1-p_k)^2N_{\ell}\times N_{\ell+1}$. Next, for analysis simplicity, we approximate the computation overhead between these two layers in the subnet to be the ratio, say $(1-p_k)^2$, of that in the original DNN for the following reason: The ratio of additions' number approaches $(1-p_k)^2$ and  the number of activation operations is far less than that of other operations. Finally, as all fully connected layers are applied with the dropout technique, the total operations to update these layers in the subnet can be approximated to $(1-p_k)^2 C_{\rm full}$ with $C_{\rm full}$ being the computation overhead to update these layers in the original DNN. Hence, the overall computational overhead can be written as
\begin{equation}\label{Eq:CMPLoad}
    C_k = C_{\rm conv} + (1-p_k)^2 C_{\rm full},
\end{equation}
where $C_{\rm conv}$ and $C_{\rm full}$ are the computational overhead for convolutional layers and fully connected layers to update the original DNN, respectively.

\subsubsection{Optimization of FedDrop rates}
By substituting the $C^2$ overhead in \eqref{Eq:Parameters} and \eqref{Eq:CMPLoad} into the per-round latency in \eqref{Eq:SystemLatency}, the minimum FedDrop rate of device $k$ can be derived as
\begin{equation}\label{Eq:MimimumRates}
p_k^{\min} = 1- \sqrt{ \dfrac{T - T_k^{\rm conv} }{ T_k^{\rm full} } },\quad 1\leq k\leq K,
\end{equation}
where $T_k^{\rm conv}$ and $ T_k^{\rm full}$ are the sum $\text{C}^2$ latencies for device $k$ to update all convolutional layers and all fully connected layers in the original DNN, respectively. They are given by 
\begin{equation}\label{Eq:LatencyConvFull}
\left\{
\begin{aligned} 
  &T_k^{\rm conv} =   \dfrac{M_{\rm conv}}{B_k} \left( \dfrac{1}{R_k^{d \ell}} +  \dfrac{1}{R_k^{u \ell}} \right) + \dfrac{C_{\rm conv} D_k}{ f_k},\\
  & T_k^{\rm full} = \dfrac{M_{\rm full}}{B_k} \left( \dfrac{1}{R_k^{d \ell}} +  \dfrac{1}{R_k^{u \ell}} \right) + \dfrac{C_{\rm full} D_k}{ f_k},
\end{aligned}
\right.
\end{equation}
where the notations follow that in \eqref{Eq:COMLatency}, \eqref{Eq:CMPLatency}, \eqref{Eq:Parameters}, and \eqref{Eq:CMPLoad}. One can observe from \eqref{Eq:MimimumRates} and \eqref{Eq:LatencyConvFull} that the dropout rate of a device is a monotone decreasing function of the associated communication rate and computation speed.

\section{Experiments and Performance Analysis}
%\subsection{Experiments Settings}
Consider a single-cell network in a disk with a radius of $0.15$ kilometres. There are one server (AP) located at the center and $K=10$ devices.  The devices are uniformly distributed in the cell. The path loss between devices and the AP is $128.1+37.6\log d $ with the distance $d$ in kilometre.  Rayleigh fading is assumed. Each device is assumed to have $B=1$ MHz bandwidth for uploading and downloading. The devices' computation capacities are uniformly selected from the set $\{0.1, 0.2, \cdots, 1.0\}$ GHz. Two \emph{convolutional neural networks} (CNNs) are trained using different settings, described as follows.
\begin{itemize}
    \item \emph{CNNCifar}: There are six convolutional layers and four fully connected layers. A pooling layer is followed after the 2rd, 4th, and 6th convolutional layers. The numbers of parameters in  convolutional layers and fully connected layers are $7,776$ and $74,000,960$, respectively. It is trained on the Cifar-10 dataset with complex features. %The global dataset includes $30,000$ samples. %The learning rate is set as $0.05$ and the optimizer is SGD.

    \item \emph{CNNMnist}: There are two convolutional layers and two fully connected layers. A pooling layer is followed after each convolutional layer. The numbers of parameters in  convolutional layers and fully connected layers are $750$ and $16,500$, respectively.  It is trained on the MNIST dataset with simple features. %The global dataset includes $30,000$ samples. %The learning rate is set as $0.01$. The optimizer is SGD.
\end{itemize}

Besides FedDrop, two benchmarking  schemes are considered. The first is the conventional  FL without dropout. The other is called \emph{uniform dropout}, where in each round, one single subnet is generated by the server by dropout and broadcast to  all devices for updating. The dropout rate of each device is the same and should be the largest: 
%\begin{equation}\label{Eq:UniformRates}
   $ p = \max\limits_k\; p_k^{\min}$,
%\end{equation}
where $p_k^{\min}$ is defined in \eqref{Eq:MimimumRates}.

\subsubsection{Learning Performance}
\begin{figure}[t]
\centering
\includegraphics[width=0.42\textwidth]{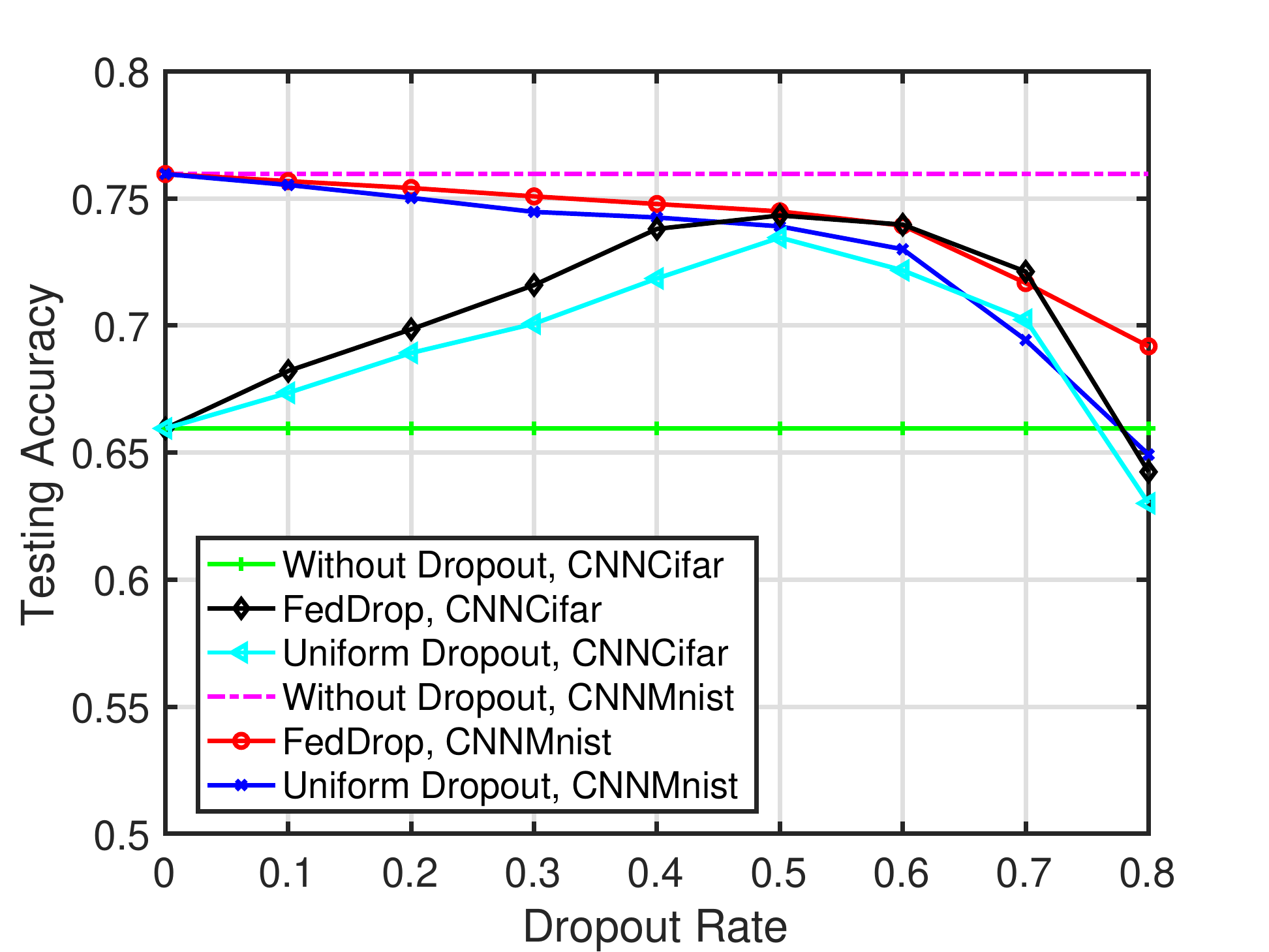}
\caption{Testing accuracy versus dropout rate.}\label{Fig:Performance}
\end{figure}

The influence of dropout rate on the learning performance, say testing accuracy after convergence, is investigated, as shown in Fig. \ref{Fig:Performance}. The two  schemes are compared by training the two models mentioned before. For fair comparison and explicitly showing the influence of dropout rate, all devices are set to have the same computation speed and spectrum efficiencies, and are assigned with same bandwidths. The channel is assumed to be static. Hence, all devices have the same dropout rate during the whole learning process. And the $C^2$ overhead of each device to update the fully connected layers in both schemes is reduced to $(1-p)^2$ with $p$ being the dropout rate.

First,  consider the case where the complex model, say CNNCifar, is trained on non \emph{independent and identical distributed} (IID) local datasets. From Fig. \ref{Fig:Performance}, the testing accuracy of both schemes first increases and then decreases,  as the dropout rate increases. The reasons are in the following. On the one hand, by applying the dropout technique, both schemes can enjoy a model diversity to avoid overfitting during training. On the other hand, high dropout rate  is harmful for the model expression ability. Besides, in Fig. \ref{Fig:Performance}, it can be observed that  the FedDrop scheme outperforms the uniform dropout scheme. The reasons are two-fold. One is that the parameters of the whole model are updated in the former instead of only a subnet in the latter. The other is that FedDrop has a greater model diversity gain by training different subnets in each round. 

Then, consider the case for training the simple model, say CNNMnist, on non IID local datasets. From Fig. \ref{Fig:Performance}, similarly, the FedDrop scheme outperforms the uniform dropout scheme for similar reasons with the previous case. Besides, we can observe that the testing accuracy of both schemes decreases with increasing dropout rate. That's because the training is underfitting, i.e., a simple model is trained on the dataset with simple features, and  the dropout technique degrades the expression ability of the CNNMnist model. However, the performance degradation is slight in both schemes, when the dropout rate is not large, i.e., $p\leq 0.6$.

From the experiments above, it can be summarized that the learning performance can benefit from the dropout technique when the training is overfitting. Moreover, the FedDrop scheme can enjoy a better model diversity than the uniform dropout scheme and thus has a better performance.

\subsubsection{$C^2$ Overhead Comparison}
\begin{figure}[t]
\centering
\includegraphics[width=0.4\textwidth]{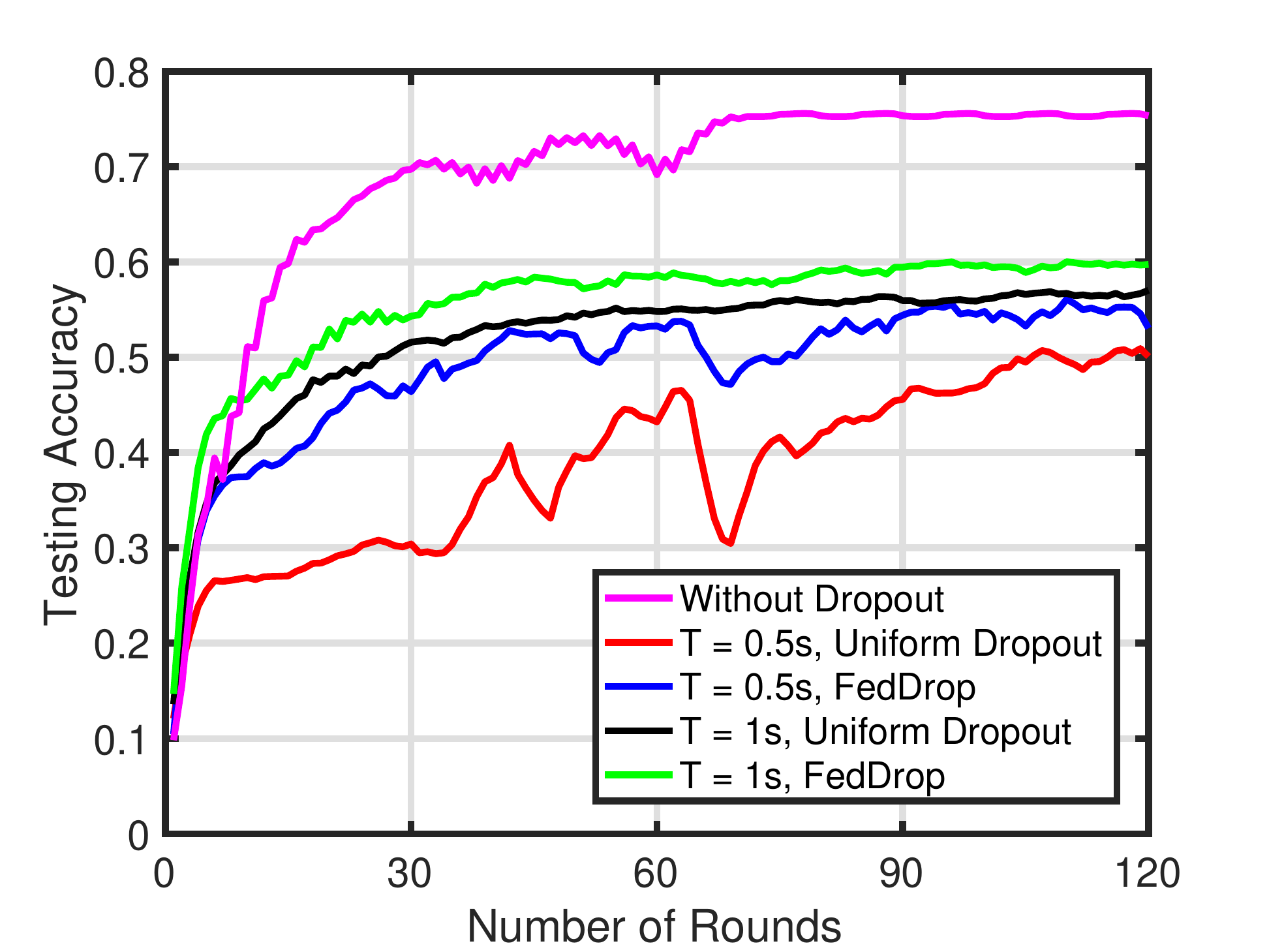}
\caption{Comparison of $C^2$ overhead between different  schemes via test their ac curacies   versus  versus the  number of rounds for varying  per-round latency constraint $T$.  }\label{Fig:C2}
\end{figure}

To compare the $C^2$ overhead of the two schemes in a more general case, the learning performance of training the CNNMnist model is compared when using the same $C^2$ resources. The schemes are compared for training the model, say CNNMnist, on the non-IID local datasets under the same per-round latency $T$, as shown in Fig. \ref{Fig:C2}.  First of all,  FedDrop always outperforms uniform dropout, demonstraing the effectiveness of the former design.  Moreover, allowing longer per-round latency improves the two schemes with dropout. With lower latency, the performance gap between the two schemes becomes larger. On the other hand, one can observe that in this case without the issue of overfitting, the $C^2$ overhead reduction of FedDropout and uniform dropout is   at the cost of some performance loss w.r.t. to FL without dropout.

\section{Concluding Remarks}
In this paper, an efficient FL algorithm  is proposed that intelligently integrates  the dropout technique to alleviate the resource constraints of edge devices. The resultant FedDrop scheme reduces the $C^2$ overhead and also acquires  the ability of coping with model overfitting. This work opens several research directions. One is to quantify the tradeoff between training latency and testing accuracy by adjusting the dropout rate. Another is to jointly design the radio resource management and load balancing in FedDrop systems.

\bibliographystyle{ieeetr}
\bibliography{reference}

\end{document}